\pdfoutput=1

\documentclass[11pt]{article}

\usepackage{naacl2021}

\usepackage{times}
\usepackage{latexsym}
\usepackage{amsmath,amssymb,amsfonts}
\usepackage{algorithmic}
\usepackage{graphicx}
\usepackage{caption}
\usepackage{subcaption}
\usepackage{textcomp}
\usepackage{xcolor}
\usepackage{hhline}
\usepackage{comment}
\usepackage{multirow,multicol}
\usepackage[ruled,vlined]{algorithm2e}
\usepackage[T1]{fontenc}

\usepackage[utf8]{inputenc}
\usepackage{booktabs}
\usepackage{xcolor}
\usepackage{tabularx}

\usepackage{microtype}
\usepackage{bbm}
\title{Improving BERT Model Using Contrastive Learning for Biomedical Relation Extraction}

\author{Peng Su\textsuperscript{\dag}, Yifan Peng\textsuperscript{\ddag,1},  K. Vijay-Shanker\textsuperscript{\dag,1}\\ 
\textsuperscript{\dag} Department of Computer and Information Science, University of Delaware \\
\textsuperscript{\ddag} Department of Population Health Sciences, Weill Cornell Medicine  \\
\texttt{\{psu, vijay\}@udel.edu}, \texttt{yip4002@med.cornell.edu} }

\begin{document}
\maketitle
\addtocounter{footnote}{1}
\footnotetext{These authors contributed equally.}

\begin{abstract}
Contrastive learning has been used to learn a high-quality representation of the image in computer vision. However, contrastive learning is not widely utilized in natural language processing due to the lack of a general method of data augmentation for text data. In this work, we explore the method of employing contrastive learning to improve the text representation from the BERT model for relation extraction. The key knob of our framework is a unique contrastive pre-training step tailored for the relation extraction tasks by seamlessly integrating linguistic knowledge into the data augmentation. Furthermore, we investigate how large-scale data constructed from the external knowledge bases can enhance the generality of contrastive pre-training of BERT. The experimental results on three relation extraction benchmark datasets demonstrate that our method can improve the BERT model representation and achieve state-of-the-art performance. In addition, we explore the interpretability of models by showing that BERT with contrastive pre-training relies more on rationales for prediction. Our code and data are publicly available at: \url{https://github.com/udel-biotm-lab/BERT-CLRE}.
\end{abstract}

\section{Introduction}

Contrastive learning is a family of methods to learn a discriminative model by comparing input pairs \cite{le-khac2020contrastive}. The comparison is performed between positive pairs of “similar” inputs and negative pairs of “dissimilar” inputs. The positive pairs can be generated in an automatic way by transforming the original data to variants without changing the key information (e.g., rotate an image). Contrastive learning can encode general properties (e.g. invariance) in the learned representation while it is relatively hard for other representation learning methods to achieve \cite{bengio2013representation, le-khac2020contrastive}. Therefore, contrastive learning provides a powerful approach to learn representations in a self-supervised manner and has shown great promise and achieved the state of the art results in recent years \cite{he2020momentum, chen2020simple}.


Despite its advancement, contrastive learning has not been well studied in biomedical natural language processing (BioNLP), especially for relation extraction (RE) tasks. One obstacle lies in the discrete characteristics of text data. Compared to computer vision, it is more challenging to design a general and efficient data augmentation method to construct positive pairs. Instead, there have been significant advances in the development of pre-trained language models to facilitate downstream BioNLP tasks \cite{devlin2018bert, radford2019language, peng2019transfer}. Therefore, leveraging contrastive learning in the large pre-trained language models to learn more general representation for RE tasks remains unexplored.


To bridge this gap, this paper presents an innovative method of contrastive pre-training to improve the language model representation for biomedical relation extraction. As the main difference from the existing contrastive learning framework, we augment the datasets for RE tasks by randomly changing the words that do not affect the relation expression. Here, we hypothesize that the shortest dependency path (SDP) between two entities \cite{bunescu2005shortest} captures the required knowledge for the relation expression. We hence keep words on SDP fixed during the data augmentation. In addition, we utilize external knowledge bases to construct more data to make the learned representation generalize better, which is a method that is frequently used in distant supervision \cite{mintz2009distant, peng2016improving}. 

To verify the effectiveness of the proposed method, we use the transformer-based BERT model as a backbone \cite{devlin2018bert} and evaluate our method on three widely studied RE tasks in the biomedical domain: the chemical-protein interactions (ChemProt) \cite{krallinger2017overview}, the drug-drug interactions (DDI) \cite{herrero2013ddi}, and the protein-protein interactions (PPI) \cite{krallinger2008overview}. The experimental results show that our method boosts the BERT model performance and achieves state-of-the-art results on all three tasks. 

Interest has also grown in designing interpretable BioNLP models that are both plausible (accurate) and rely on a specific part of the input (faithful rationales) \cite{deyoung2019eraser, lei2016rationalizing}. Here rationale is defined as the supporting evidence in the inputs for the model to make correct predictions. In this direction, we propose a new metric, ''prediction shift'', to measure the sensitivity degree to which the small changes (out of the SDP) of the inputs will make model change its predictions. We show that the contrastively pre-trained model is more robust than the original model, suggesting that our model is more likely to make predictions based on the rationales of the inputs. 

In sum, the contribution of this work is four-fold. (1) We propose a new method that utilizes contrastive learning to improve the BERT model on biomedical relation extraction tasks. (2) We utilize external knowledge to generate more data for learning more generalized text representation. (3) We achieve state-of-the-art performance on three benchmark datasets of relation extraction tasks. (4) We propose a new metric that aims to reveal the rationales that the model uses for predicting relations. The code and the new rationale test datasets are available at \url{https://github.com/udel-biotm-lab/BERT-CLRE}.

\section{Related Work}

The history of contrastive representation learning can be traced back to \cite{hadsell2006dimensionality}, in which the authors explore the method of representation learning that similar inputs are mapped to nearby points in the representation space. Recently, with the development of data augmentation techniques, deep neural network architectures, contrastive learning regains attention and achieves superior performance on visual representation learning \cite{he2020momentum, chen2020simple}. In \cite{he2020momentum}, the Momentum Contrast (MoCo) framework is designed to learn representation using the mechanism of dictionary look-up: an encoded example (the query) should be similar to its matching key (augmented sample from the same data example) and dissimilar to others. In \cite{chen2020simple}, the authors propose the SimCLR frame to learn the representations by maximizing the agreement between augmented views of the same data point.

The contrastive representation has all the properties that a good representation should have: 1) Distributed property; 2) Abstraction and invariant property; 3) Disentangled representation \cite{bengio2013representation, le-khac2020contrastive}. The distributed property emphasizes the expressive aspect of the representation (different data points should have distinguishable representations). The capture of abstract concepts and the invariance to small and local changes are concerned in the abstraction and invariant property. From the disentangled representation's perspective, it should encode as much information as possible. In this work, we will show contrastive learning can improve the invariant aspect of the representation.

In the natural language processing (NLP) field, several works have utilized the contrastive learning technique. \citet{fang2020cert} propose a pre-trained language representation model (CERT) using contrastive learning at the sentence level to benefit the language understanding tasks. \citet{klein2020contrastive} employ contrastive self-supervised learning to solve the commonsense reasoning problem. \citet{peng2020learning} propose a self-supervised pre-training framework for relation extraction to explore the encoded information for the textual context and entity type. Compared with the previous works, we employ different data augmentation techniques and utilize data from external knowledge bases in contrastive learning to improve the model for relation extraction tasks.

Relation extraction is usually seen as a classification problem when the entity mentions are given in the text. Many different methods have been proposed to solve the relation extraction problem \cite{culotta2004dependency, sierra2008definitional, sahu2018drug, zhang2019deep, su2019using}. However, the language model methods re-define this field with their superior performance \cite{dai2015semi, peters2018deep, devlin2018bert, radford2019language, su2020investigation}. Among all the language models, BERT \cite{devlin2018bert} --a language representation model based on bidirectional Transformer \cite{vaswani2017attention}, attracts lots of attention in different fields. Several BERT models have been adapted for biomedical domain: BioBERT \cite{lee2020biobert}, SciBERT \cite{beltagy2019scibert}, BlueBERT \cite{peng2019transfer} and PubMedBERT \cite{gu2020domain}. BioBERT, SciBERT and BlueBERT are pre-trained based on the general-domain BERT using different pre-training data. In contrast, PubMedBERT \cite{gu2020domain} is pre-trained from scratch using PubMed abstracts.

In recent years, there is increasing interest in designing more interpretable
NLP models that reveal the logic behind model predictions. In \cite{deyoung2019eraser}, multiple datasets of rationales (from human experts) are collected to facilitate the research on interpretable models in NLP. In \cite{lei2016rationalizing}, the authors propose an encoder-generator framework to automatically generate candidate rationales to justify the predictions of neural network models.

\section{Methodology}

\subsection{The framework of contrastive learning}
\begin{figure}[tb]
\centering
\includegraphics[width=0.45\textwidth]{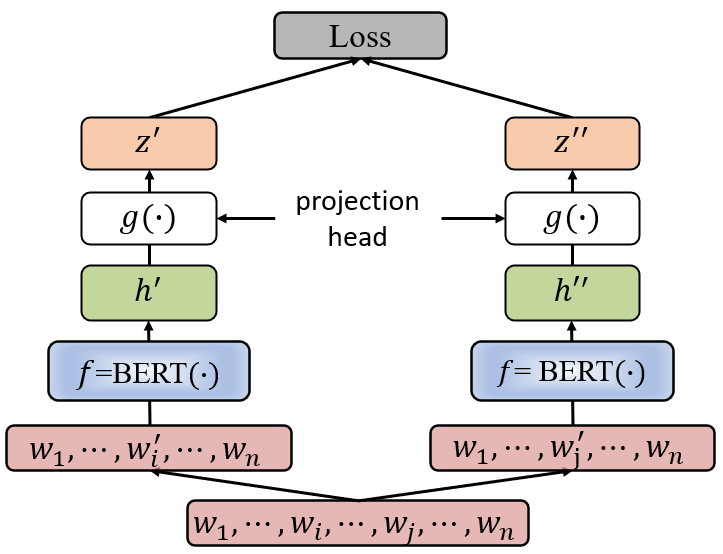}
\caption{\label{fig:frame} The framework of contrastive learning. For the data augmentation of relation extraction, we randomly replace some words that are not affecting the relation expression ($w_i\rightarrow w_i^{'}$ in the left sample, $w_j\rightarrow w_j^{'}$ in the right sample).}
\end{figure}

Our goal is to learn a text representation by maximizing agreement between inputs from positive pairs via a contrastive loss in the latent space and the learned representation can then be used for relation extraction. Figure \ref{fig:frame} shows our framework of contrastive learning. Given a sentence $s = w_1,...w_n$, we first produce two augmented views (a positive pair) $v'=w_1,...,w_i',...w_n$ and $v''=w_1...,w_j',...w_n$ ($i \neq j$) from $s$ by applying text augmentation technique (Section~\ref{sec:augmentation}). 

Our framework then uses one neural network to encode the two inputs, which consists of a neural network encoder $f$ (Section~\ref{sec:encoder}) and a projection head $g$ (Section~\ref{sec:projector}). From the first augmented view $v'$, we output a \textit{representation} $h'\triangleq f(v')$ and a projection $z'\triangleq g(h')$. From the second augmented view $v''$, we output $h''\triangleq f(v'')$ and another projection $z''\triangleq g(h'')$.

The contrastive learning method learns the representation by comparing different samples in the training data (Section~\ref{sec:loss}). The comparison is performed between both similar inputs and dissimilar inputs, and the similar inputs are positive pairs and the dissimilar inputs are negative pairs. During the training, the representations are learned by leading the positive pairs to have similar representations and making negative pairs have dissimilar representations. In applications, the positive pairs are usually from the augmented data of the same sample, and the negative pairs are generated by selecting augmented data from different samples.

At the end of training, we only keep the encoder $f$ as in \cite{chen2020simple}. For any text input $x$, $h=f(x)$ will be the representation of $x$ from contrastive learning.


\begin{table*}
\centering
\begin{tabular}{ll}
\toprule
Original&We further show that \underline{@PROTEIN\$} directly \underline{interacts} with \underline{@PROTEIN\$} and Rpn4.\\
After SR& We further show that \underline{@PROTEIN\$} \textbf{straight} \underline{interacts} with \underline{@PROTEIN\$} and Rpn4.\\
After RS& \textbf{Further we} show that \underline{@PROTEIN\$} directly \underline{interacts} with \underline{@PROTEIN\$} and Rpn4.\\
After RD& We further show that \underline{@PROTEIN\$} \underline{interacts} with \underline{@PROTEIN\$} and Rpn4.\\
\bottomrule
\end{tabular}
\caption{\label{augmentation}Examples after the three operations for data augmentation. The shortest dependency path between two proteins is "@PROTEIN\$ interacts @PROTEIN\$", which is marked with underline in the examples. The changed words are also marked with bold font.}
\end{table*}

\subsubsection{Data augmentation for relation extraction}
\label{sec:augmentation}

The data augmentation module is a key component of contrastive learning, which needs to randomly generate two correlated views for the original data point. At the same time, the generated data should be different from each other to make them distinguishable (from the model's perspective), but should not be significantly different to change the structure and semantics of the original data. It is especially difficult to augment the text data of relation extraction. In this work, we only focus on binary relations. 
Given $<s,e_1,e_2,r>$, where $e_1$ and $e_2$ are two entity mentions in the sentence $s$ with the relation type $r$, we keep $e_1$ and $e_2$ in the sentence and retain the relation expression between $e_1$ and $e_2$ in the augmented views.

Specifically, we propose a data augmentation method utilizing the shortest dependency path (SDP) between the two entities in the text. We hypothesize that the shortest dependency path captures the required information to assert the relationship of the two entities \cite{bunescu2005shortest}. Therefore we fix the shortest dependency path, and randomly change the other tokens in the text to generate the augmented data. This idea is inspired by \cite{wei2019eda}, which employed easy data augmentation techniques to improve model performance on text classification tasks.

As the preliminary study, we experiment with three techniques to randomly replace the tokens to generate the augmented data and choose the best one for our contrastive learning method: 1) Synonym replacement (SR), 2) Random swap (RS), and 3) Random deletion (RD). 

Table \ref{augmentation} gives some samples after applying the three operations on a sentence from the PPI task. For the synonym replacement, we randomly replace $n$ words with their synonyms. To acquire the synonym of a word, we utilize the WordNet database \cite{miller1995wordnet} to extract a list of synonyms and randomly choose one from the list. For the random swap, we swap the positions of two words and repeat this operation $n$ times. For the random deletion, we delete some words with the probability $p$. The probability $p$ is set to 0.1 in our experiments and the parameter $n$ for SR and RS is calculated by $p\times l$, where $l$ is the length of the sentence.

To examine which operation performs better for relation extraction tasks, we train three BERT models using the three types of augmented data (combined with the original training data). Table \ref{aug_performance} shows that the synonym replacement (SR) operation achieves the best performance on all three tasks and we will employ this operation in our data augmentation module in our contrastive learning experiments (We will further discuss it in Section~\ref{sec:dataaugmentation}).

\subsubsection{The neural network encoder}
\label{sec:encoder}

In this work, we employ the BERT model \cite{devlin2018bert} as our encoder for the text data and the classification token ([CLS]) output in the last layer will be the representation of the input. 

\subsubsection{Projection head}
\label{sec:projector}

As demonstrated in \cite{chen2020simple}, adding a nonlinear projection head on the model output will improve the representation quality during training. Following the same idea, a multi-layer perceptron (MLP) will be applied to the model output $h$. Formally, $$z=g(h)=W^2 \phi(W^1h)$$ and $\phi$ is the ReLU activation function, $W^1$ and $W^2$ are the weights of the perceptron in the hidden layers.

\subsubsection{Contrastive loss}
\label{sec:loss}

Contrastive learning is designed to make similar representations be learned for the augmented samples (positive pairs) from the same data point. We follow the work of \cite{chen2020simple} to design the loss function (Algorithm \ref{algo}). During contrastive learning, the contrastive loss is calculated based on the augmented batch derived from the original batch. Given $N$ sentences in a batch, we first employ the data augmentation technique to acquire two views for each sentence in the batch. Therefore, we have $2N$ views from the batch. Given one positive pair (two views from the same sentence), we treat the other $2(N-1)$ within the batch as negative examples. Similar to \cite{chen2020simple},
the loss for a positive pair is defined as:
$$l(z',z'')=-log \frac{exp(sim(z',z'')/\tau)}{\sum_{k=1}^{2N} \mathbbm{1}_{[z_k\ne z']}  exp(sim(z',z_k)/\tau)}$$
where $sim(\cdot,\cdot)$ is the cosine similarity function, $\mathbbm{1}_{[z_k\ne z']}$ is the indicator function and $\tau$ is the temperature parameter. The final loss $L$ is computed across all positive pairs, both $(z',z'')$ and ($z''$, $z'$), in a batch.

For computation convenience, we arrange the $(2k-1)$-th example and the $2k$-th example in the batch are generated from the same sentence, a.k.a., $(2k-1, 2k)$ is a positive pair. Please see Algorithm \ref{algo} for calculating the contrastive loss in one batch. Then we can update the parameters of the BERT model and projection head $g$ to minimize the loss $L$. 

\begin{algorithm}
\SetAlgoLined
 Input: encoder $f$ (BERT), project head $g$, data augmentation module, data batch $\{s_k\}_{k=1}^N$\;
 \For{k=1,...,N}{
  $v',v'' = data\_augment(s_k)$\;
  $z_{2k-1}=g(f(v'))$\;
  $z_{2k}=g(f(v''))$\;
 }
 $L=\frac{1}{2N} \sum_{k=1}^N[l(z_{2k-1},z_{2k})+l(z_{2k},z_{2k-1})]$
 \caption{\label{algo}Contrastive loss in a batch}
\end{algorithm}

\subsection{Training procedure}
\begin{figure}[tb]
\centering
\includegraphics[width=0.45\textwidth]{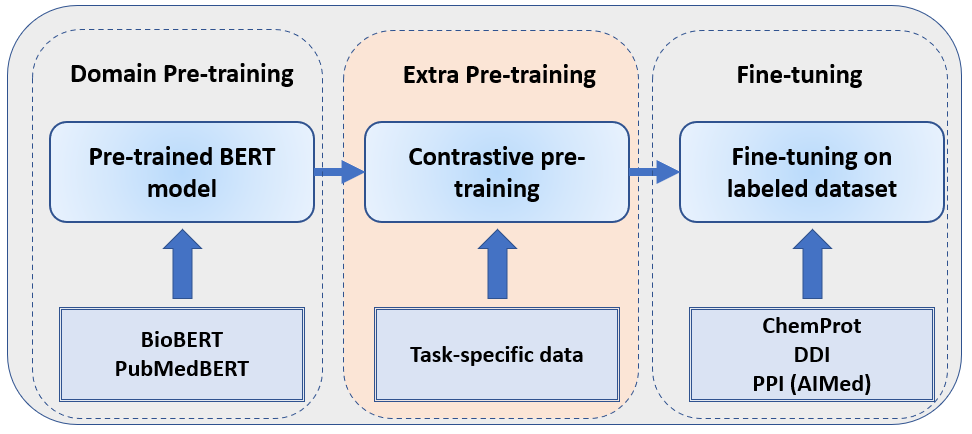}
\caption{\label{fig:pretraining} The pipeline of BERT model training with contrastive pre-training.}
\end{figure}

Figure \ref{fig:pretraining} shows the training procedure of our framework. It consists of three stages. First, we pre-train the BERT model on a large amount of unlabeled data from a specific domain(e.g., biomedical domain). Second, we conduct contrastive pre-training on task-specific data as a continual pre-training step after the domain pre-training of BERT model. In this way, we retain the learned knowledge from general pre-training, and add the new features from contrastive learning. Finally, we fine-tune the model on the RE tasks to further gain task-specific knowledge through supervised training on the labeled datasets.

The domain pre-training stage follows that of the BERT using the masked language model and next sentence prediction technique \cite{devlin2018bert}. In our experiments, we use two pre-trained versions for the biomedical domain: BioBERT \cite{lee2020biobert} and PubMedBERT \cite{gu2020domain}.

\subsection{A knowledge-based method to enrich training dataset for contrastive learning}

Contrastive pre-training requires a large-scale dataset to generalize the representation. Also, our data augmentation for contrastive learning needs SDP between two given entities, so we need to construct the augmented dataset with the entities mentioned in the text. For these purposes, we utilize external databases for the relations to acquire extra instances for contrastive learning.

Formally, assuming a curated database for relation $r$ contains all the relevant entities and text, we consider every combination of the entity pairs in one sentence and use them as examples for this relation. For instance, there are three proteins in the sentence $s$: "Thus \underline{NIPP1} works as a molecular sensor for \underline{PP1} to recognize phosphorylated \underline{Sap155}." We will generate three examples for PPI task from this sentence: <$s$,NIPP1,PP1,PPI>, <$s$,NIPP1,Sap155,PPI> and <$s$,PP1,Sap155,PPI>. 

We use the IntAct database \cite{orchard2014mintact} as the interacting protein pairs database for the PPI task. Similarly, DrugBank \cite{wishart2008drugbank} and BioGRID \cite{stark2006biogrid} are utilized for DDI and ChemProt, respectively. In the column "EK" of Table \ref{tab:statistics}, we show the statistics of datasets for each task generated by external knowledge bases. We can see that the datasets from the external database are much larger than that of the human-labeled datasets. 

\begin{table}
\centering
\begin{tabular}{@{}lrrrr@{}}
\toprule
Task & Train & Dev & Test & EK \\
\midrule
ChemProt & 18,035&11,268& 15,745 & 35,500 \\
DDI & 22,233 & 5,559&5,716 & 67,959\\
PPI$^*$ &  5,251 & - & 583 & 97,853 \\
\bottomrule
\end{tabular}
\caption{Statistics of datasets used for contrastive pre-training and fine-tuning. EK: datasets generated by external knowledge bases; *: since there is no standard split of training and test set for the PPI dataset (AIMed), we use 10-fold cross-validation and here we show number of the training and test in each fold.}
\label{tab:statistics}
\end{table}

\begin{table*}
\centering
\begin{tabular}{l c c c  c c c  c c c }
\toprule
\multirow{2}{*}{Model}&\multicolumn{3}{c}{ChemProt}&\multicolumn{3}{c}{DDI} & \multicolumn{3}{c}{PPI}\\
\cmidrule(lr){2-4}\cmidrule(lr){5-7}\cmidrule(lr){8-10}
&  P & R & F & P & R & F & P & R & F \\
\midrule
BioBERT & 74.3 & \textbf{76.3} & 75.3 &79.9 & 78.1 & 79.0 &79.0 & \textbf{83.3} & 81.0 \\
BioBERT+CL & \textbf{77.0} & 74.7 & 75.8 &82.6 & 77.4 & 79.9&79.8 & 83.1 & 81.3 \\
BioBERT+CLEK & 76.6 & 76.0 & \textbf{76.3}& \textbf{82.9} & \textbf{78.4} & \textbf{80.6} & \textbf{81.1} & 83.2& \textbf{82.1} \\
\midrule
PubMedBERT & 78.8 & 75.9 & 77.3 &82.6 & 81.9 & 82.3&\textbf{80.1} & 84.3 & 82.1 \\
PubMedBERT+CL & 79.6 & 76.2 & 77.8 &\textbf{83.3} & 81.5 & 82.4&79.4 & 85.6 & 82.4 \\
PubMedBERT+CLEK & \textbf{80.6} & \textbf{76.9} & \textbf{78.7} & \textbf{83.3} & \textbf{82.4} & \textbf{82.9} & 79.9 & \textbf{85.7}& \textbf{82.7} \\
\bottomrule
\end{tabular}
\caption{BERT model performance on ChemProt, DDI and PPI tasks. BioBERT/PubMedBERT: original BERT model; BioBERT/PubMedBERT+CL: BioBERT/PubMedBERT with contrastive pre-training on the training set of human-labeled dataset; BioBERT/PubMedBERT+CLEK: BioBERT/PubMedBERT with contrastive pre-training on the data from the external knowledge base. }
\label{tab:biobert}
\end{table*}

\section{Experiments}

As discussed before, we will utilize the BERT model as the encoder for the inputs. In particular, we will employ two BERT models pre-trained for the biomedical domain in our experiments: BioBERT \cite{lee2020biobert} and PubMedBERT \cite{gu2020domain}. 

\subsection{Datasets and evaluation metrics}

We will evaluate our method on three benchmark datasets. The statistics of these datasets is shown in Table~\ref{tab:statistics}. For ChemProt and DDI tasks, we employ the corpora in \cite{krallinger2017overview} and \cite{herrero2013ddi} respectively, and we use the same split of training, development and test sets with the PubMedBERT model \cite{gu2020domain} during the model evaluation. We utilize the AIMed corpus \cite{bunescu2005comparative} for the PPI task, and we will employ 10-fold cross-validation on it since there is no standard split of training and test. 

PPI is a binary classification problem, and we will use the standard precision (P), recall (R) and F1-score (F) to measure the model performance. However, the ChemProt and DDI tasks are multi-class classification problems. The ChemProt corpus is labeled with five positive classes and the negative class: CPR:3, CPR:4, CPR:5, CPR:6, CPR:9 and negative. Similar to the DDI corpus, there are four positive labels and one negative label: ADVICE, EFFECT, INT, MECHANISM and negative. The models for ChemProt and DDI will be evaluated utilizing micro precision, recall and F1 score on the non-negative classes.

\subsection{Data pre-processing}

One instance of relation extraction task contains two parts: the text and the entity mentions. In order to make the BERT model identify the positions of the entities, we replace the relevant entity names with predefined tags by following the standard pre-processing step for relation extraction \cite{devlin2018bert}. Specifically, all the protein names are replaced with @PROTEIN\$, drug names with @DRUG\$, and chemical names with @CHEMICAL\$. In Table \ref{augmentation}, we show a pre-processed example of the PPI task. 

\subsection{Training setup}

For the fine-tuning of the BioBERT models, we use the learning rate of 2e-5, batch size of 16, training epoch of 10, and max sequence length of 128. During the fine-tuning of PubMedBERT models, the learning rate of 2e-5, batch size of 8, training epoch of 10 and max sequence length of 256 are utilized. 

In the contrastive pre-training step of the BERT models, we use the same learning rate with the fine-tuning, and the training epoch is selected from [2, 4, 6, 8, 10] based on the performance on the development set. If there is no development set (e.g., PPI task), we will use 6 as the default training epoch. Since contrastive learning benefits more from larger batch \cite{chen2020simple}, we utilize the batch size of 256 and 128 for BioBERT and PubMedBERT respectively. In addition, the temperature parameter $\tau$ is set to 0.1 during the training.

\section{Results and discussion}

\subsection{BERT model performance with contrastive pre-training}
Table \ref{tab:biobert} demonstrates the experimental results using the BERT models with contrastive pre-training and external datasets. The first row is the BioBERT model performance without applying contrastive learning. The following two rows demonstrate the results after adding the contrastive pre-training step in BioBERT. The "BioBERT+CL" stands for the BioBERT model with contrastive pre-training on the training set of the human-labeled dataset, while "BioBERT+CLEK" is for the BioBERT model with contrastive pre-training on the data from the external knowledge base. Similarly, we give the PubMedBERT model performance of our method in the last three rows of Table \ref{tab:biobert}. 

We can see that the contrastive per-training improves the model performance in both cases. However, contrastive pre-training on human-labeled dataset only improves the model with a small margin. We hypothesize that the limited improvement might be due to the poor generalization on small training set. Therefore, we include more data (EK data) in contrastive learning to enhance the model generalizability. The data generated from the external knowledge base are much more than the training data of the human-labeled dataset (column "EK" and "train" in Table \ref{tab:statistics}). As shown in the third and sixth row in Table \ref{tab:biobert}, contrastive learning with more external data can further boost the model performance. Compared with the BERT models without contrastive pre-training, we observe an averaged F1 score improvement (on the two BERT models) of 1.2\%, 1.2\%, and 0.85\% on ChemProt, DDI, and PPI datasets, respectively.  

Since PubMedBERT is the state-of-the-art (SOTA) model on these three tasks, we further improve its performance by adding contrastive pre-training. Thus, we achieve state-of-the-art performance on all three datasets.

\subsection{Comparison of data augmentation techniques}
\label{sec:dataaugmentation}
\begin{table}
\centering
\begin{tabular}{lrrr}
\toprule
Training data& ChemProt&DDI& PPI\\
\midrule
Original&75.3&79.0&81.0\\
+RS&75.6&78.4&75.4\\
+RD&75.4&79.8&81.2\\
+SR&\textbf{76.0}&\textbf{80.1}&\textbf{81.9}\\
\bottomrule
\end{tabular}
\caption{\label{aug_performance} BioBERT model performance (F1 score) using different types of augmented data. RS: random swap; RD: random deletion; SR: synonym replacement.}
\end{table}

Table \ref{aug_performance} shows the BERT model performance after including three types of augmented data. We can see that the synonym replacement (SR) operation yields the best results on all three tasks. Therefore we use it as our default operation to generate augmented data in all our contrastive learning experiments. We also notice that the augmented data from the random swap (RS) operation hurt the model performance on the DDI and PPI tasks, which indicates that this operation might change the relation expression in the sentence. Thus it is necessary to verify the effectiveness of the operations before applying them on contrastive learning.

\subsection{Measurement of rationale faithfulness}

\begin{table}
\centering
\small
\begin{tabular}{@{}l@{~}p{18em}@{~}r@{}}
\toprule
& Input sentence & Prediction \\
\midrule
(1) & Instead, radiolabeled @CHEMICAL\$ resulting from @PROTEIN\$ hydrolysis were observed. & CPR:9\\
(2) & \textbf{Or else}, radiolabeled @CHEMICAL\$ resulting from @PROTEIN\$ hydrolysis were observed. & False\\
\midrule
(1) & These results indicate that membrane @PROTEIN\$ levels in N-38 neurons are dynamically autoregulated by @CHEMICAL\$. & CPR:3\\
(2) & These results indicate that membrane @PROTEIN\$ levels in N-38 \textbf{nerve cell} are dynamically autoregulated by @CHEMICAL\$. & False\\
\bottomrule
\end{tabular}
\caption{\label{example}Examples of prediction shift. (1): Original sentence; (2): Augmented sentence.}
\end{table}

\begin{table}[h]
\centering
\begin{tabular}{@{}l@{~~}lr@{~}c@{}}
\toprule
Task & Model & \multicolumn{2}{c}{Prediction} \\
& & \multicolumn{2}{c}{Shift}\\
\midrule

\multirow{4}{*}{ChemProt}  & BioBERT& 246 & \\
 & BioBERT+CLEK& 191 & (22\% $\downarrow$)\\
 \cline{2-4}
& PubMedBERT& 248 & \\
 & PubMedBERT+CLEK& 189 & (24\% $\downarrow$)\\
\midrule

\multirow{4}{*}{DDI} & BioBERT& 111 & \\
 & BioBERT+CLEK& 89 & (20\% $\downarrow$)\\
 \cline{2-4}
& PubMedBERT& 90 & \\
 & PubMedBERT+CLEK& 75 & (17\% $\downarrow$)\\
\midrule
\multirow{4}{*}{PPI$^{*}$} & BioBERT& 51 & \\
 & BioBERT+CLEK& 33 & (35\%$\downarrow$)\\
 \cline{2-4}
& PubMedBERT& 49 & \\
 & PubMedBERT+CLEK& 34 & (31\%$\downarrow$)\\
\bottomrule
\end{tabular}
\caption{Count of prediction shift on the "augmented" test set. *: The sum of counts on the 10 folds.}
\label{tab:invariance}
\end{table}

As discussed previously, we hypothesize the words on the shortest dependency path (SDP) as the rationales in the input. Therefore, the model should make its predictions based on them. If the model predictions are all made based on a specific part of the input, we can define this specific part of the input to be the completely faithful rationales. In practice, the rationales are more faithful means they are more influential on the model predictions.    

In this work, we define a new metric to measure the faithfulness of the rationales: "prediction shift". If the model predicts one test example (non-negative) with label $L_t$, but changes its prediction on its neighbor (the augmented data point) with another label $L_t^{'}$, we will say a "prediction shift" happens (In Table \ref{example}, we give two examples of prediction shift on PubMedBERT model). Fewer "prediction shift" indicates the information outside of SDP influences the prediction less, which means the rationales are more faithful.

To generate a similar set (with test set) for the measurement of "prediction shift", we apply the same synonym replacement (SR) technique on the original test data. Since we retain the words that are on the shortest dependency path between the two entities, the generated data should express the same relation with the original ones. The trained model should predict them with the same labels if the rationales of input are utilized during inference, and in that case, we say the rationales are faithful.

We compare the number of "prediction shift" on two types of BERT model: the original BERT and the BERT model with contrastive pre-training. Table \ref{tab:invariance} illustrates that the BERT models with contrastive pre-training dramatically reduce the number of "prediction shift". Those results indicate that the BERT models with contrastive pre-training rely more on the information of shortest dependency path for prediction, a.k.a., the rationales are more faithful. From another perspective, the results in Table \ref{tab:invariance} also demonstrate that the BERT models with contrastive pre-training are resilient to small changes of the inputs, which means the models are more robust. 

\section{Conclusion and Future Directions}
In this work, we propose a contrastive pre-training method to improve the text representation of the BERT model. Our approach differs from previous studies in the choice of text data augmentation with linguistic knowledge and the use of the external knowledge bases to construct large-scale data to facilitate contrastive learning. The experimental results demonstrate that our method outperforms the original BERT model on three relation extraction benchmarks. Additionally, our method shows robustness to slightly changed inputs over the BERT models. In the future, we will investigate different settings of data augmentation and contrastive pre-training to exploit their capability on language models. We also hope that our work can inspire researchers to design better metrics and create high-quality datasets for the exploration of model interpretability.

\section*{Acknowledgment}

Yifan Peng's research was supported by National Library of Medicine - National Institutes of Health (NIH) under award number R00LM013001. Peng Su's graduate studies were supported by a supplement of NIH grant U01GM125267.

\bibliography{anthology,custom}
\bibliographystyle{acl_natbib}

\appendix

\end{document}